\documentclass{article}

\newcommand{\V}[1]{\mathbf{#1}}
\usepackage[accepted]{icml2019}

\usepackage{float}
\usepackage{amsmath,bm}
\usepackage{subcaption}
\usepackage{caption}
\usepackage{color}
\usepackage{appendix}
\usepackage{subcaption}
\usepackage[utf8]{inputenc}
\usepackage{graphicx,bm}
\usepackage{amsfonts}
\usepackage{bbold}
\usepackage{url}

\usepackage[
    left = ``,%
    right = '',%
    leftsub = `,%
    rightsub = '%
]{dirtytalk}

\graphicspath{{fig/}{fig/8features/}{fig/4features}{fig/8features/GBT}{fig/8features/RF}}
\icmltitlerunning{Understanding Uncertainty in LIME Explanations}

\begin{document}

\twocolumn[
\icmltitle{\say{Why Should You Trust My Explanation?} \\
Understanding Uncertainty in LIME Explanations}




\begin{icmlauthorlist}
\icmlauthor{Yujia Zhang$^*$}{cor}
\icmlauthor{Kuangyan Song$^*$}{h}
\icmlauthor{Yiming Sun$^*$}{cor}
\icmlauthor{Sarah Tan}{cor}
\icmlauthor{Madeleine Udell}{cor}
\end{icmlauthorlist}
\icmlaffiliation{cor}{Cornell University}
\icmlaffiliation{h}{Zshield Inc}
\icmlcorrespondingauthor{Yujia Zhang}{yz685@cornell.edu}
\icmlkeywords{interpretability, explanations, uncertainty, LIME}
\vskip 0.3in
]
\printAffiliationsAndNotice{\icmlEqualContribution} 
\begin{abstract}
Methods for explaining black-box machine learning models aim to increase the transparency of these model and provide insights into the reliability and fairness of such models. However, the explanations themselves could contain significant uncertainty that undermines users' trust in the predictions and raises concern about the model's robustness. Focusing on a particular local explanations method, Local Interpretable Model-Agnostic Explanations (LIME), we demonstrate the presence of three sources of uncertainty, namely randomness in the sampling procedure, variation with sampling proximity, and variation in explained model credibility across different data points. Such uncertainty is present even for black-box models with high test accuracy. We investigate the uncertainty in the LIME method on synthetic data and two public data sets, newsgroups text classification and recidivism risk-scoring.

\end{abstract}

\section{Introduction}

While machine learning models have become increasingly important for decision making in many areas \cite{zeng2016interpretable,rajkomar2018scalable}, many machine learning models are “black-box” in that the process by which such models make predictions can be hard for humans to understand. Explanations of model predictions can help increase users' trust in the model \cite{lipton2016mythos,Ribeiro}, determine if the model achieves desirable properties such as fairness, privacy, etc. \cite{Doshi2017}, and debug possible errors in the model \cite{ribeiro2018semantically}.

Indeed, explanation methods aim to help users assess and establish trust in black-box models and their predictions. However, whether the explanations themselves are trustworthy is not obvious. Uncertainty in explanations not only cast doubt on the understanding of a certain prediction, but also raises concerns about the reliability of the black-box model in the first place, hence diminishing the value of the explanation \cite{Fragile2017}.


In this paper, we address the question: when can we trust an explanation? In particular, we study the local explanation method Local Interpretable Model-Agnostic Explanations (LIME) \cite{Ribeiro}. Briefly, LIME explains the prediction of a desired input by sampling its neighboring inputs and learning a sparse linear model based on the predictions of these neighbors; features with large coefficients in the linear model are then considered to be important for that input's prediction. We demonstrate that training LIME explanations involve sources of uncertainty that should not be overlooked. More specifically, generating a local explanation for an input requires sampling around the input to generate an explanation for its prediction. In this paper, we show that this sampling can lead to statistical uncertainty in interpretation.  

\section{Related Work}
The study of interpretable methods can be roughly divided into two fields -- designing accurate, yet still inherently interpretable models \cite{letham2015,lakkaraju2016interpretable}, and creating post-hoc methods to explain black-box models, either locally around a specific input \cite{baehrens2010explain,Ribeiro} or globally for the entire model \cite{ribeiro2018anchors,tan2018transparent}. In this paper, we study one particular local explanation method, LIME \cite{Ribeiro}.

Several sensitivity-based explanation methods for neural networks \cite{shrikumar2017learning,selvaraju2017grad,Sundararajan2017IntGrad} have been shown to be fragile \cite{Fragile2017,adebayo2018sanity}. \citeauthor{Fragile2017} demonstrated that it is possible to generate vastly different explanations for two perceptively indistinguishable inputs with the same predicted labels from the neural network. This paper focuses on the fragility of local post-hoc explanations of models.

Potential issues with LIME's stability and robustness have been pointed out by \citeauthor{Alvarez2018robust}, who showed that while LIME explanations can be stable when explaining linear models, for nonlinear models this is not always the case \cite{Alvarez2018robust}. Testing LIME on images, \citeauthor{Lee2019sensitivity} observed that  LIME colored superpixels differently across different iterations and proposed an aggregated visualization to reduce the perception of different explanations over different iterations \cite{Lee2019sensitivity}. However, they did not study the source of this instability of explanations -- the focus of our paper. 

\section{Approach}
\subsection{Uncertainty in LIME Explanations}
Given a black box model $f$, and a target 
point $x$ to be explained, LIME samples neighbors of $x$ and their black-box outcomes and chooses a model $g$ from some interpretable functional space $G$ by solving 
\begin{equation}
\text{argmin}_{g\in G} \mathcal{L}(f, g, \pi_x) + \Omega(g)
\end{equation}
where $\pi_x$ is some probability distribution around $x$ and $\Omega(g)$ is a penalty for model complexity. \citeauthor{Ribeiro} \cite{Ribeiro} suggests several methods to achieve sparse solution, including K-LASSO as the interpretable model. For K-LASSO, we let $\Omega = \infty \mathbb{1}[\|w_g\|_0 > K]$, where $w$ denotes the coefficients of the linear model, and sample points near $x$ from $\pi_x$ to train K-LASSO. We observe that this procedure involves three sources of uncertainty:
\begin{itemize}
\item Sampling variance in explaining a single data point;
\item Sensitivity to choice of parameters, such as sample size and sampling proximity;
\item Variation in explanation on model credibility across different data points.
\end{itemize}

\subsection{Methodology}
We use one synthetic data example and two real datasets to demonstrate the three aforementioned sources of uncertainty. To show the sampling variance, we run LIME multiple times for a single data point, record the top few features selected by K-LASSO each time, and observe the cumulative selection probability for each selected feature. Whether features are consistently selected over different trials reflects LIME's instability in explaining the data point. Then, we tune the parameters of LIME to probe the sensitivity of the explanations to sample size and sampling proximity. Finally, we compare LIME explanations of different data points by assessing whether the selected features are informative in the real context. Variation in explanation on model credibility across different data points raises concern about the credibility of LIME as a global explanation for the model. 

\section{Results}
We first use synthetic tree-generated data to illustrate the first and second source of uncertainty mentioned above. Then we use examples in text classification to demonstrate the third source of uncertainty. Finally we apply LIME to the COMPAS dataset as a case where LIME explanations are considered trustworthy.
\subsection{Synthetic data generated by trees}

\begin{figure}
\centering
\includegraphics[width=0.5\textwidth]{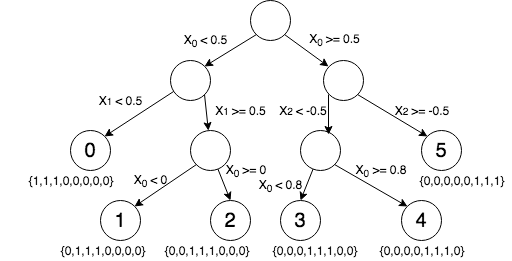}
\caption{Decision tree partition of eight-feature synthetic data. The local coefficients on each leaf are shown under each end node. Data points are assigned labels based on the linear classifier in Equation \ref{eq:assign-label}.}
\label{fig:8-generate-graph}
\end{figure}

\textbf{Data:} Given the number of features $N$, we generate training and test data from local sparse linear models on uniformly distributed input in $[0,1]^N$. To illustrate LIME's local behavior at different data points, we partition them with a known decision tree. Within each partition, we assign labels on each data point $\V{x}$ based on a linear classifier with known coefficients $\bm \beta$ as shown in Equation \ref{eq:assign-label}. 
\begin{equation}
\label{eq:assign-label}
y(\V{x}) = \left\{
\begin{array}{cc}
    1 &  \V{x}^\top \bm\beta  \ge 0\\
    0 &  \V{x}^\top \bm\beta  < 0.
\end{array} \right.
\end{equation}
We consider two cases where the number of features is 4 and 8 respectively. Figure \ref{fig:8-generate-graph} presents a way of splitting the data into six leaves for $N=8$ with known coefficients, where three out of eight features have coefficients 1 in each leaf. 
The data splitting and coefficients for $N=4$ are presented in Figure \ref{fig:assign-tree-4-features} in Appendix \ref{sec: more-simu-setting}. 


\begin{figure*}
  \centering
  \begin{subfigure}[b]{0.3\textwidth}
    \centering
    \includegraphics[width=\linewidth]{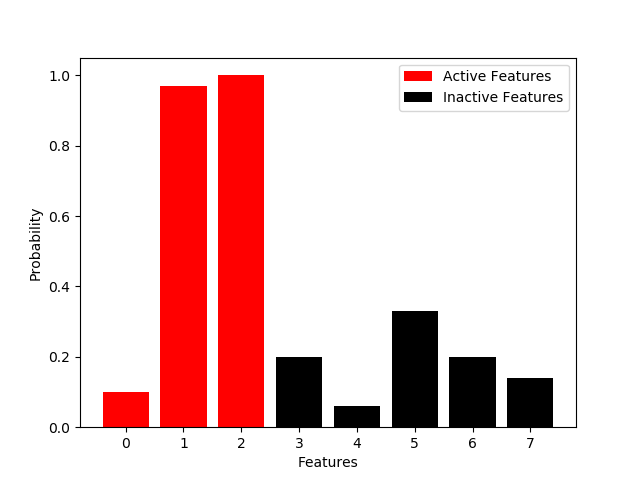}
    \caption{Leaf 0}
    \label{fig:8-label0}
  \end{subfigure}
  ~
  \begin{subfigure}[b]{0.3\textwidth}
    \centering
    \includegraphics[width=\linewidth]{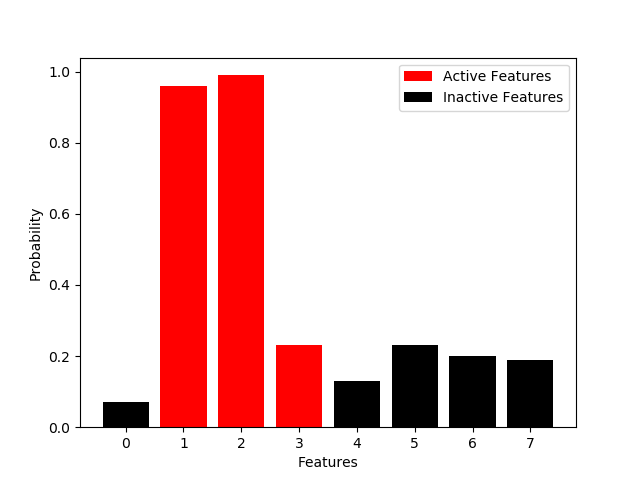}
    \caption{Leaf 1}
    \label{fig:8-label1}
  \end{subfigure}
  ~
  \begin{subfigure}[b]{0.3\textwidth}
    \centering
    \includegraphics[width=\linewidth]{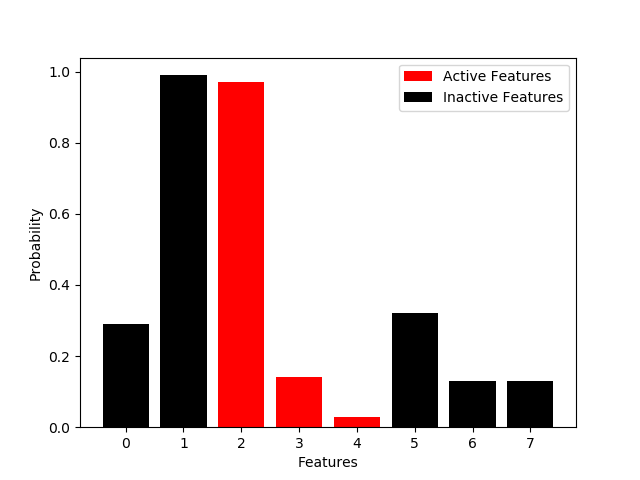}
    \caption{Leaf 2}
    \label{fig:8-label2}
  \end{subfigure}

  \begin{subfigure}[b]{0.3\textwidth}
    \centering
    \includegraphics[width=\linewidth]{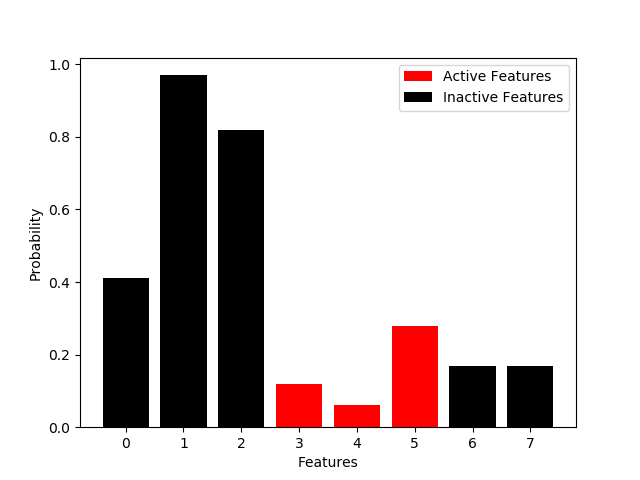}
    \caption{Leaf 3}
    \label{fig:8-label3}
  \end{subfigure}
  ~
  \begin{subfigure}[b]{0.3\textwidth}
    \centering
    \includegraphics[width=\linewidth]{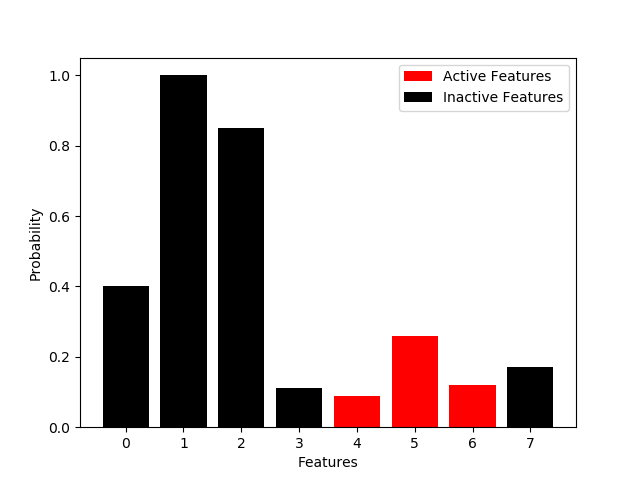}
    \caption{Leaf 4}
    \label{fig:8-label4}
  \end{subfigure}
  ~
  \begin{subfigure}[b]{0.3\textwidth}
    \centering
    \includegraphics[width=\linewidth]{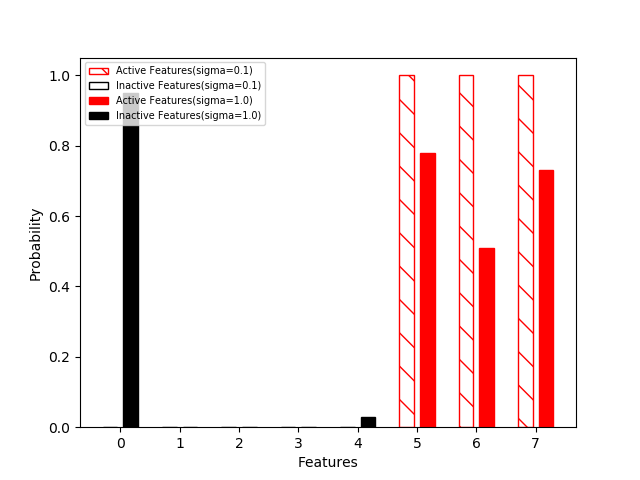}
    \caption{Leaf 5}
    \label{fig:8-label5}
  \end{subfigure}
  \caption{Empirical selection probability in LIME explanations of the random forest model trained by eight-feature synthetic data. A data point is taken from each leaf, and LIME is run 100 times on each point. For each trial of LIME, we record the three features selected by K-Lasso, and calculate the cumulative selection probability for each of the eight features. For each leaf, active features with true coefficients 1 are marked red. Notice that features chosen by LIME are not necessarily locally important features on each leaf. Especially for leaves 3-5, signal from the true features is dominated by signal from the first three features used for tree splitting. For leaf 5, we also tried reducing the sampling proximity by a factor of ten (striped bars), which allows us to recover significant signal from the true local features and rule out the signal of feature 0 used for splitting.}
  \label{fig:8-feature-RF}
\end{figure*}

\textbf{Results:} We present results for the case where we apply LIME to interpret black-box models (random forest and gradient boosting tree) trained with eight-feature synthetic data. We run LIME on one data point in each of the six leaves. We first notice that different trials potentially select different features due to sampling variance. Figure~\ref{fig:8-feature-RF} shows the cumulative selection frequency of top three features in each trial when LIME interprets a random forest model; the case with gradient boosting is shown in Figure~\ref{fig:8-feature-GBT} in Appendix \ref{sec: more-simu-setting}. LIME captures the signal of the first three features, which are used globally in the tree splitting of the data. Locally, however, different features are important for each individual leaf, which LIME fails to reflect. Thus, its explanation cannot be considered stable around each input data point in a tree structure. We further notice that LIME by default draws samples from a rescaled standard normal distribution $\mathcal{N}(0,\sigma^2)$ near the test point, where $\sigma^2$, the variance of the training data, determines the sampling proximity. The experiments show that LIME tends to capture locally important features better with a smaller sampling proximity and pick up global features with a larger sampling proximity. Since tuning this parameter allows LIME to explore both global and local structure in the data, we suggest users to think consciously about the choice of its value. As an example, we tune LIME's sampling proximity for a data point on leaf 5 in the eight-feature synthetic data, shown in Figure \ref{fig:8-label5}. When a sample is drawn from $\mathcal{N}(0,\sigma^2)$ near the test point, LIME captures the global features used for tree splitting; when a sample is drawn from $\mathcal{N}(0,{(0.1\sigma)}^2)$, LIME successfully picks up signal from the three local features 5-7. 
Results for running the same procedure on four-feature synthetic data are presented in Figures \ref{fig:random-forest-4-features} and \ref{fig:gradient-boosting-4-features} in Appendix \ref{sec: more-simu-setting}. 

\subsection{Text Classification} 
\textbf{Data:} The 20 Newsgroup dataset is a collection of ca. 20,000 news documents across 20 newsgroups. As noted in \cite{Ribeiro}, even for text classification models with high test accuracy, some feature words that LIME selects are quite arbitrary and uninformative. To examine this behavior further, we use Multinomial Naive Bayes classifier for two examples of document classification, namely \say{Atheism vs. Christianity} and \say{electronics vs. crypt}.

\begin{figure*}
  \centering
  \begin{subfigure}[b]{0.4\textwidth}
    \centering
    \includegraphics[width=\linewidth]{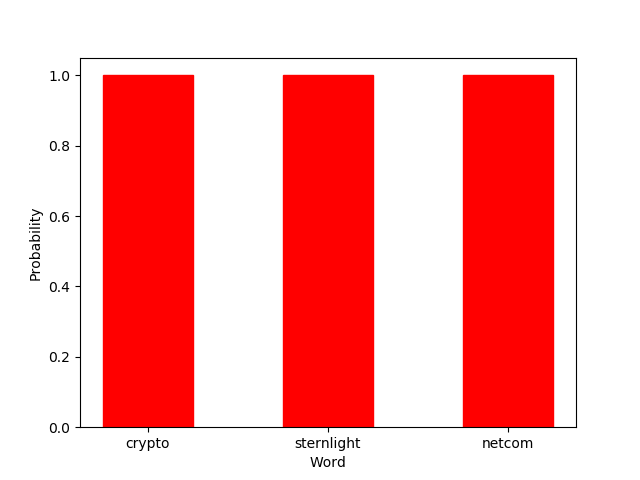}
    \caption{Test document 1}
 \label{fig:nlp-feature-counts.png}
  \end{subfigure}
  ~
  \begin{subfigure}[b]{0.4\textwidth}
    \centering
   \includegraphics[width=\linewidth]{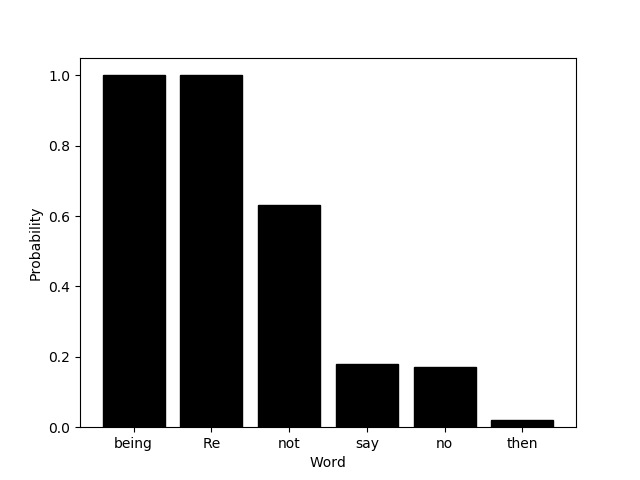}
   \caption{Test document 2}
    \label{fig:nlp-feature-counts-2.png}
  \end{subfigure}
 \caption{Empirical selection probability for feature words in text classification \say{electronics vs. crypt}. LIME is run 100 times on the test document; in each LIME trial, we record the three feature words selected by K-Lasso and calculate the empirical selection probability for these words. Words that are informative are marked red. It can be seen that the selected feature words for the first document are consistent and meaningful, while those for the second document are not informative.}
  \label{fig:mnb-nlp-2}
\end{figure*}

\textbf{Results:} Multinomial Naive Bayes classifiers are trained for the aforementioned two classification examples, with test accuracy 0.9066 and 0.9214 respectively. 
However, as pointed out in \cite{Ribeiro}, we need to know the feature importance for each output in order to establish trust in the model. In particular, we find that LIME's local explanations are not always plausible for different test documents. As shown in Figure~\ref{fig:mnb-nlp-2}, the selected feature words for the first document (\say{crypto}, \say{sternlight} and \say{netcom}) display no variation for different trials and are relevant in content, which makes the model seem very credible. However, the selected feature words for the second document are not informative at all. Thus, the model's credibility, as explained by LIME, varies across different input data. We also include results for \say{Christianity vs. Atheism} in Figure \ref{fig:mnb-nlp} in Appendix \ref{sec: more-simu-setting}, which also display a difference in model credibility for different documents. 

\subsection{COMPAS Recidivism Risk Score Dataset} 
\textbf{Data:} The \say{Correctional Offender Management Profiling for Alternative Sanctions} (COMPAS) is a risk-scoring algorithm developed by Northpointe to assess a criminal defendant's likelihood to recidivate. The risk is classified as \say{High}, \say{Medium} and \say{Low} based on crime history and category, jail time, age, demographics, etc. We study a subset of the COMPAS dataset collected and processed by ProPublica \cite{COMPAS}, with the goal of examining the presence of demographic bias in risk-scoring. As we do not have access to the true COMPAS model, we train a random forest classifier as a \say{mimic model} \cite{Tan2018distill}, using selected features and risk assessment text labels from COMPAS. We examine salient features selected by  LIME explanations on multiple COMPAS records.

\textbf{Results:} We test and analyze LIME explanation of the random forest classifier with both numerical and categorical features. Unlike the uncertainty we observe in previous experiments on synthetic and 20 Newsgroup data, we see consistent explanation results on different test data points. LIME is applied to two data points that are classified as \say{high risk} by COMPAS. The results are shown in Figure~\ref{fig:random-forest-compas} in Appendix \ref{sec: more-simu-setting}. We consider these explanations to be trustworthy due to the following two observations: 1) there is little variation in the selection of important features in different trials on the same data point, and 2) explanation is consistent for different data points, since the same features are selected for the two different data points, including race and age. Further analysis using LIME suggests that the mimic model is using demographic properties as important features in predicting a risk score. This in turn shows it is probable that the COMPAS model makes use of demographic features for recidivism risk assessment, so further investigation would be meaningful to gauge the fairness of the algorithm.


\section{Conclusion}

Explanation methods for black-box models may themselves contain uncertainty that calls into question the reliability of the black-box predictions and the models themselves. We demonstrate the presence of three sources of uncertainty in the explanation method \say{Local Interpretable Model-agnostic Explanations} (LIME), namely the randomness in its sampling procedure, variation with sampling proximity, and variation in explained model credibility for different data points. The uncertainty in LIME is illustrated by numerical experiments on synthetic data, text classification examples in 20 Newsgroup data and recidivism risk-scoring in COMPAS data.

\clearpage 
\bibliographystyle{icml2019}
\bibliography{reference.bib}
\newpage
\appendix 
\section{More Simulation Setting}
\label{sec: more-simu-setting}
\subsection{Setting for four-feature synthetic data}
\begin{figure}
\centering
\includegraphics[width=0.9\linewidth]{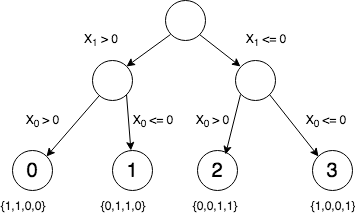}
\caption{Decision tree partition of four-feature synthetic data.}
\label{fig:assign-tree-4-features}
\end{figure}

\begin{figure*}
  \centering
  \begin{subfigure}[b]{0.3\textwidth}
    \centering
    \includegraphics[width=\linewidth]{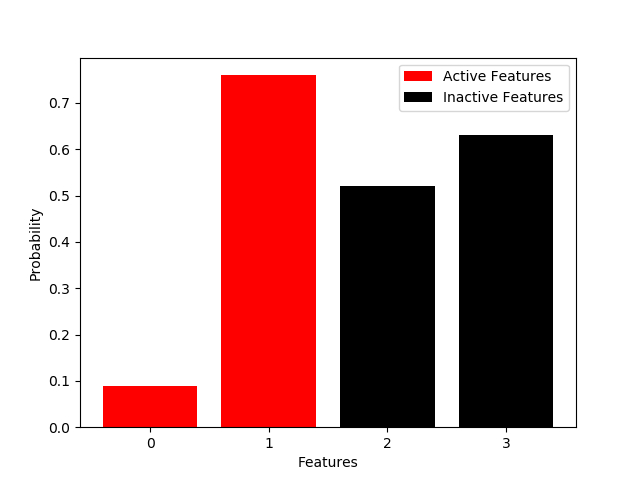}
    \caption{Leaf 0}
    \label{fig:4-label0}
  \end{subfigure}
  ~
  \begin{subfigure}[b]{0.3\textwidth}
    \centering
    \includegraphics[width=\linewidth]{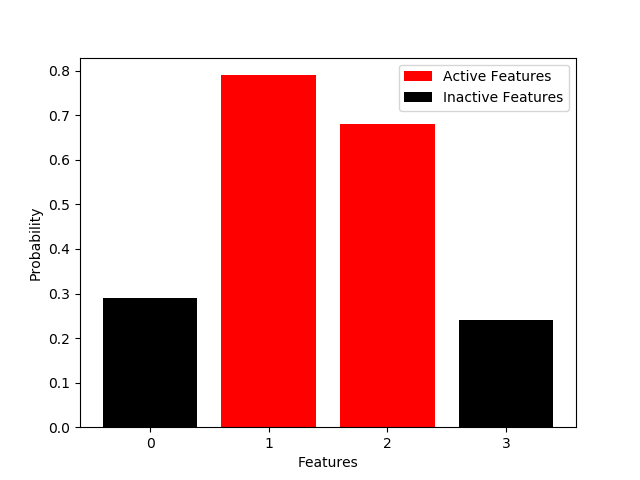}
    \caption{Leaf 1}
    \label{fig:4-label1}
  \end{subfigure}
  
  \begin{subfigure}[b]{0.3\textwidth}
    \centering
    \includegraphics[width=\linewidth]{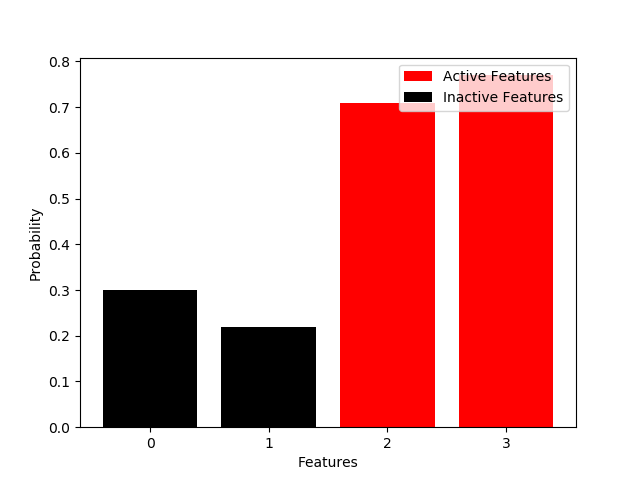}
    \caption{Leaf 2}
    \label{fig:4-label2}
  \end{subfigure}
    ~
  \begin{subfigure}[b]{0.3\textwidth}
    \centering
    \includegraphics[width=\linewidth]{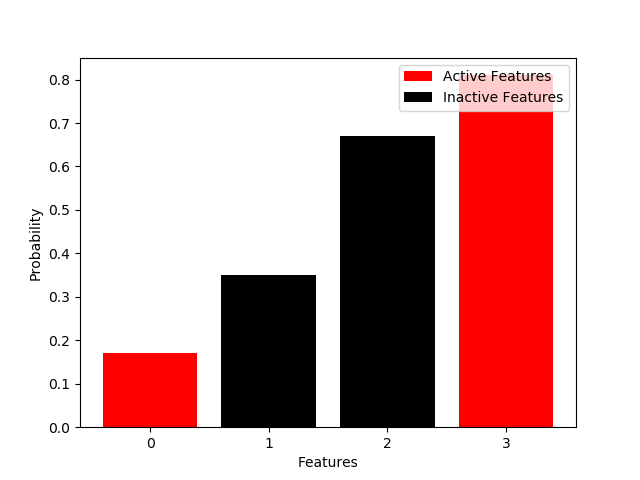}
    \caption{Leaf 3}
    \label{fig:4-label3}
  \end{subfigure}
  
  \caption{Empirical selection probability in LIME explanations of the random forest model trained by four-feature synthetic data. A data point is taken from each leaf, and LIME is run 100 times on each point. For each trial of LIME, we record the two features selected by K-LASSO, and calculate the cumulative selection probability for each of the four features. For each leaf, active features with true coefficients 1 are marked red.}
  \label{fig:random-forest-4-features}
\end{figure*}

\begin{figure*}
  \centering
  \begin{subfigure}[b]{0.3\textwidth}
    \centering
    \includegraphics[width=\linewidth]{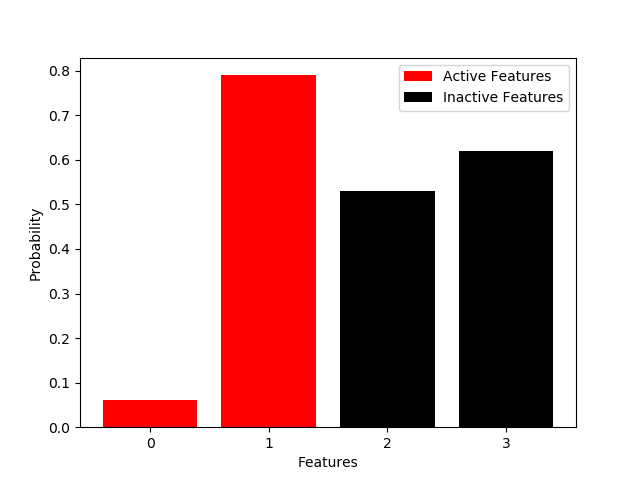}
    \caption{Leaf 0}
    \label{fig:4-label0-GBT}
  \end{subfigure}
  ~
  \begin{subfigure}[b]{0.3\textwidth}
    \centering
    \includegraphics[width=\linewidth]{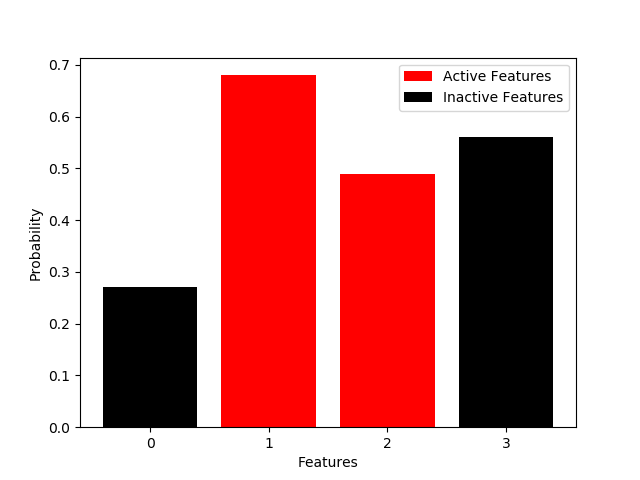}
    \caption{Leaf 1}
    \label{fig:4-label1-GBT}
  \end{subfigure}
  
  \begin{subfigure}[b]{0.3\textwidth}
    \centering
    \includegraphics[width=\linewidth]{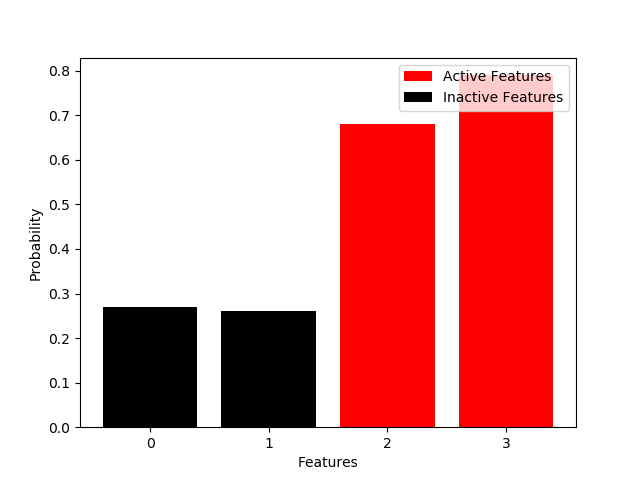}
    \caption{Leaf 2}
    \label{fig:4-label2-GBT}
  \end{subfigure}
    ~
  \begin{subfigure}[b]{0.3\textwidth}
    \centering
    \includegraphics[width=\linewidth]{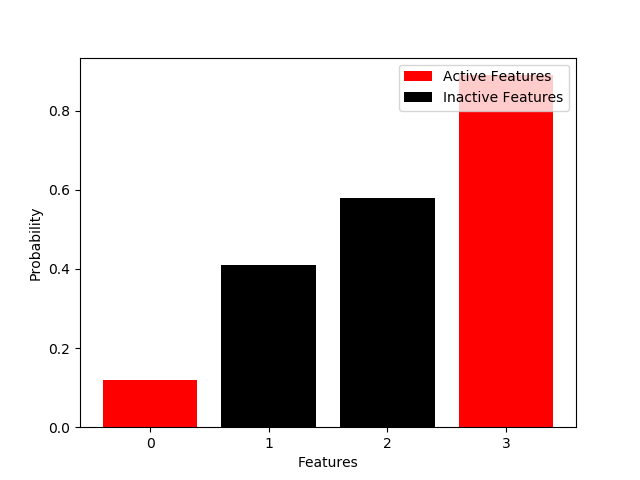}
    \caption{Leaf 3}
  \end{subfigure}
  
  \caption{Empirical selection probability in LIME explanations of the gradient boosting tree model trained by four-feature synthetic data. A data point is taken from each leaf, and LIME is run 100 times on each point. For each trial of LIME, we record the two features selected by K-LASSO, and calculate the cumulative selection probability for each of the four features. For each leaf, active features with true coefficients 1 are marked red.} 
   \label{fig:gradient-boosting-4-features}
\end{figure*}

For sample points with four features, we use the first two dimensions of features as their x and y coordinates. We assign each quadrant to different leaf, ignoring the sample points on the x or y axis. For each leaf, we assign different coefficients to their features, as shown in Figure \ref{fig:assign-tree-4-features}. We fit and explain both random forest and gradient boosting classifier within this setting, see Figure \ref{fig:random-forest-4-features} and Figure \ref{fig:gradient-boosting-4-features}.

\subsection{More Numerical Results}
\begin{figure*}
  \centering
  \begin{subfigure}[b]{0.3\textwidth}
    \centering
    \includegraphics[width=\linewidth]{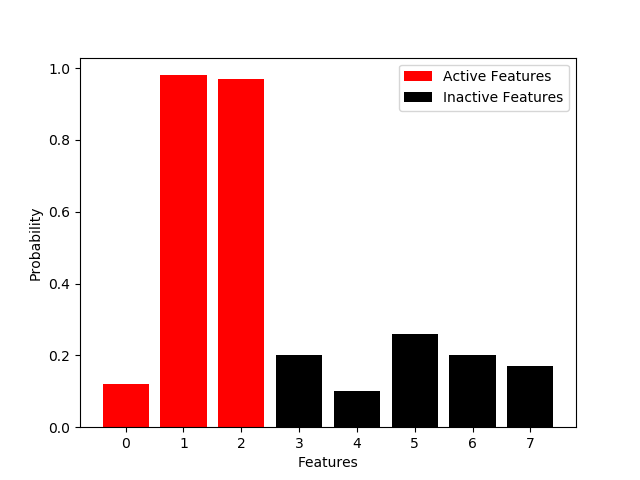}
    \caption{Leaf 0}
    \label{fig:8-label0-G}
  \end{subfigure}
  ~
  \begin{subfigure}[b]{0.3\textwidth}
    \centering
    \includegraphics[width=\linewidth]{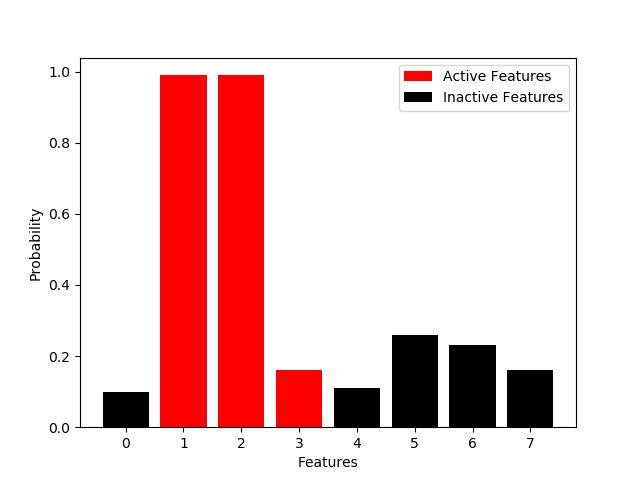}
    \caption{Leaf 1}
    \label{fig:8-label1-G}
  \end{subfigure}
  ~
  \begin{subfigure}[b]{0.3\textwidth}
    \centering
    \includegraphics[width=\linewidth]{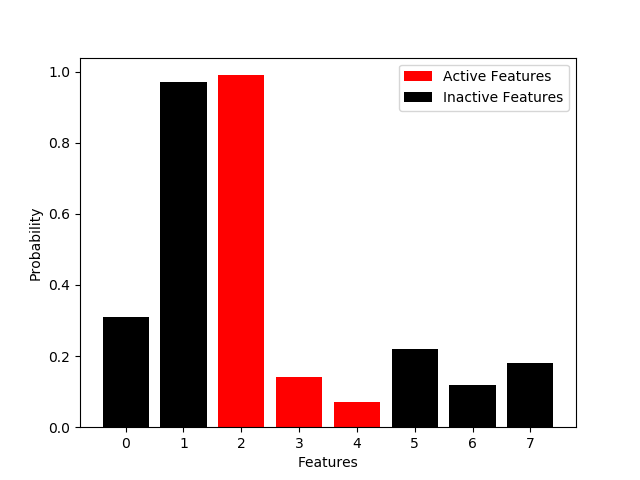}
    \caption{Leaf 2}
    \label{fig:8-label2-G}
  \end{subfigure}

  \begin{subfigure}[b]{0.3\textwidth}
    \centering
    \includegraphics[width=\linewidth]{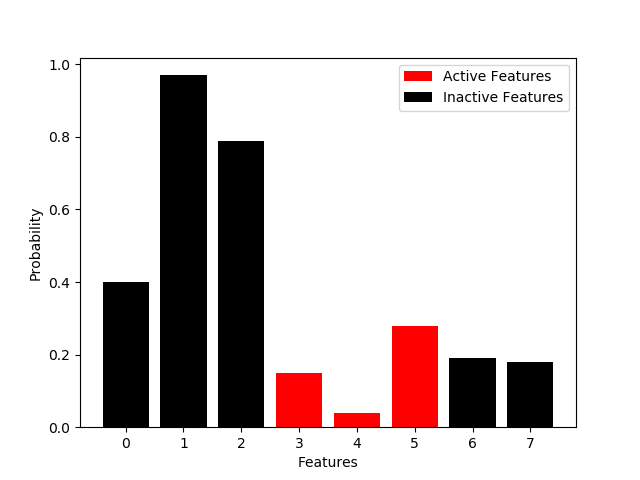}
    \caption{Leaf 3}
    \label{fig:8-label3-G}
  \end{subfigure}
  ~
  \begin{subfigure}[b]{0.3\textwidth}
    \centering
    \includegraphics[width=\linewidth]{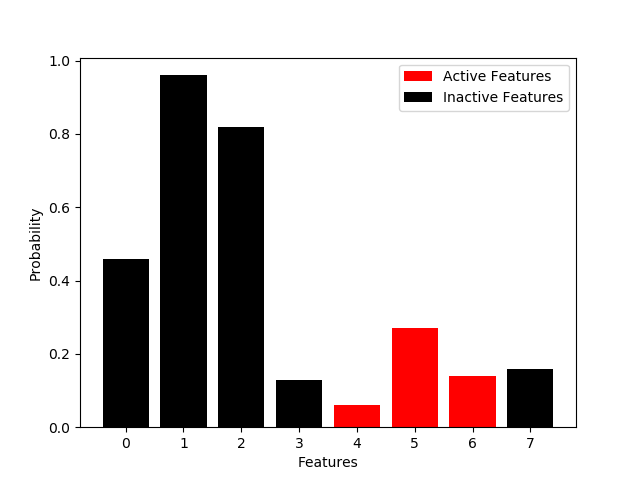}
    \caption{Leaf 4}
    \label{fig:8-label4-G}
  \end{subfigure}
  ~
  \begin{subfigure}[b]{0.3\textwidth}
    \centering
    \includegraphics[width=\linewidth]{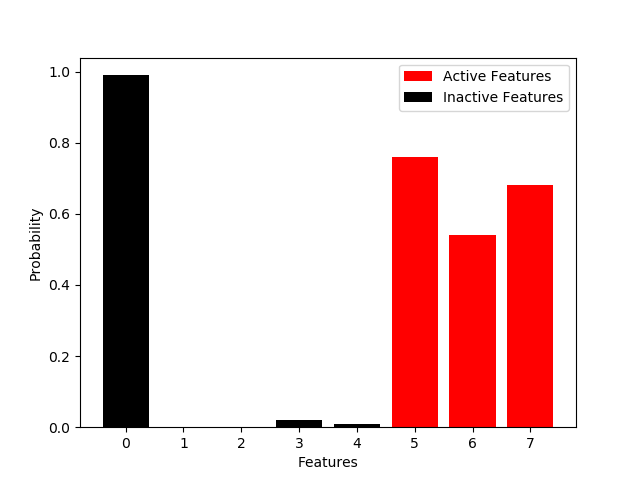}
    \caption{Leaf 5}
    \label{fig:8-label5-G}
  \end{subfigure}
  \caption{Empirical selection probability in LIME explanations of the gradient boosting tree model trained by eight-feature synthetic data. A data point is taken from each leaf, and LIME is run 100 times on each point. For each trial of LIME, we record the three features selected by K-LASSO, and calculate the cumulative selection probability for each of the eight features. For each leaf, active features with true coefficients 1 are marked red. As in the random forest model, here LIME mainly captures the global features used for tree splitting and fails to reflect the local features important on leaves 3-4. For leaf 5, we reduce LIME's sampling proximity by a factor of 10. This allows us to recover a significant amount of signal from the local features 5-7, although signal from feature 0 is still present.}
  \label{fig:8-feature-GBT}
\end{figure*}

\paragraph{Text Classification:}
We select two classes from 20 newsgroup dataset, then apply term frequency-inverse document freque (tf-idf) vectorizer with default settings. Stop words are not removed from resulting tokens as we would like to see if the model is using irrelevant features to predict the results. For the the classification between \say{Electronics} and \say{Crypt}, we analyze the explanation over two different test data points. We could see from the results that the explanation of test data document one contains several indicative words, such as \say{crypto}, \say{netcom} and \say{Sternlight} in this case. However, the explanation results for test data point two contains only one indicative word \say{information}. 
We also include example for and the result  is shown in Figure \ref{fig:mnb-nlp} in Appendix \ref{sec: more-simu-setting}.

\begin{figure*}
  \centering
  \begin{subfigure}[b]{0.49\textwidth}
    \centering
    \includegraphics[width=\linewidth]{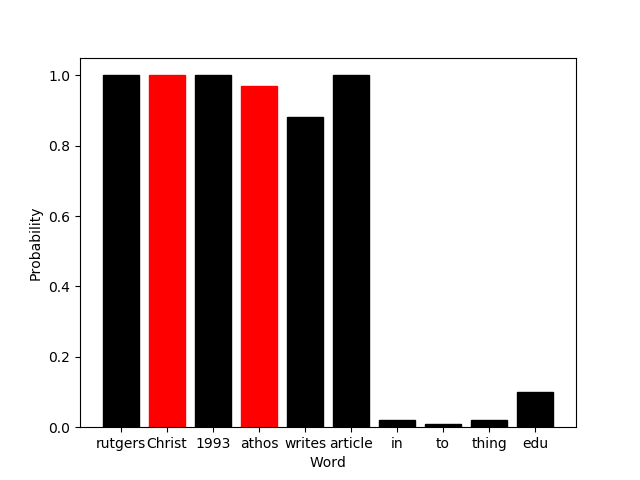}
    \caption{Test document 1}
 \label{fig:nlp-feature-counts.png}
  \end{subfigure}
  ~
  \begin{subfigure}[b]{0.49\textwidth}
    \centering
   \includegraphics[width=\linewidth]{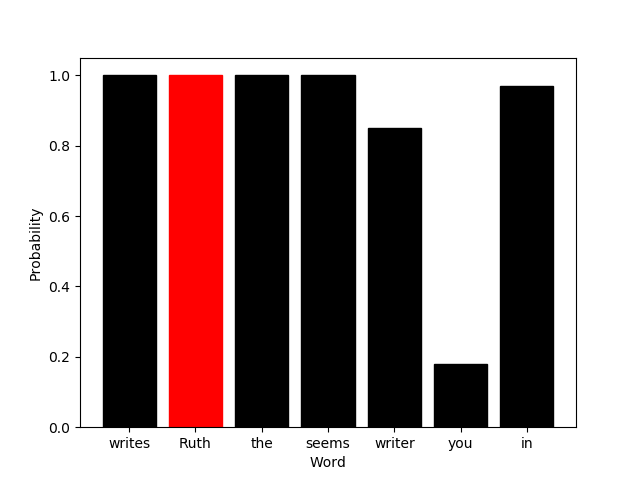}
   \caption{Test document 2}
    \label{fig:nlp-feature-counts-2.png}
  \end{subfigure}
 \caption{Empirical selection probability for feature words in text classification \say{Christianity vs. Atheism}. LIME is run 100 times on the test document; in each LIME trial, we record the six feature words selected by K-Lasso and calculate the empirical selection probability for these words. Different trials potentially select different feature words. Words that are informative are marked red. It can be seen that many of the frequently selected feature words are not informative.}
  \label{fig:mnb-nlp}
\end{figure*}

\paragraph{COMPAS Recidivism Risk Score Data:}
The COMPAS dataset from ProPublica contains a lot irrelevant columns, as well as null values. We selected twelve relevant columns of the dataset, then drop the rows that contain null value. Specifically, we exclude the decile score columns as it is directly related to the text label. We then encode the categorical features, such as \say{sex} and \say{race}, using one-hot encoder, and encode the label text using label encoder. After the simple data pre-process, we trained a random forest classifier on the processed dataset to mimic the COMPAS black-box model, which we do not have access to.
\begin{figure*}
  \centering
  \begin{subfigure}[b]{0.49\textwidth}
    \centering
    \includegraphics[width=\linewidth]{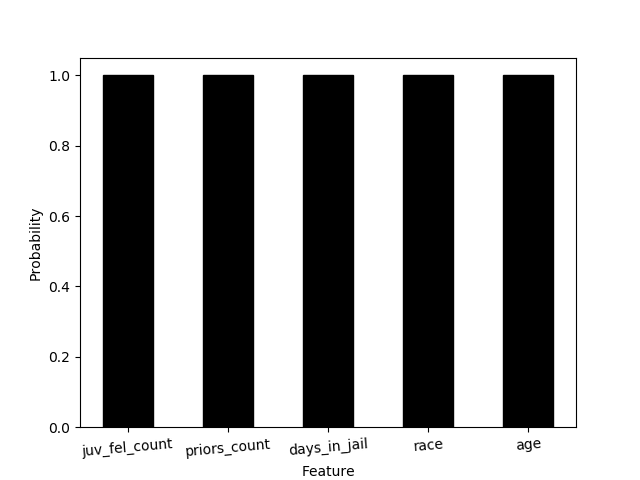}
    \caption{Sample data 1}
     \label{fig:compas-1.png}
  \end{subfigure}
  ~
  \begin{subfigure}[b]{0.49\textwidth}
    \centering
    \includegraphics[width=\linewidth]{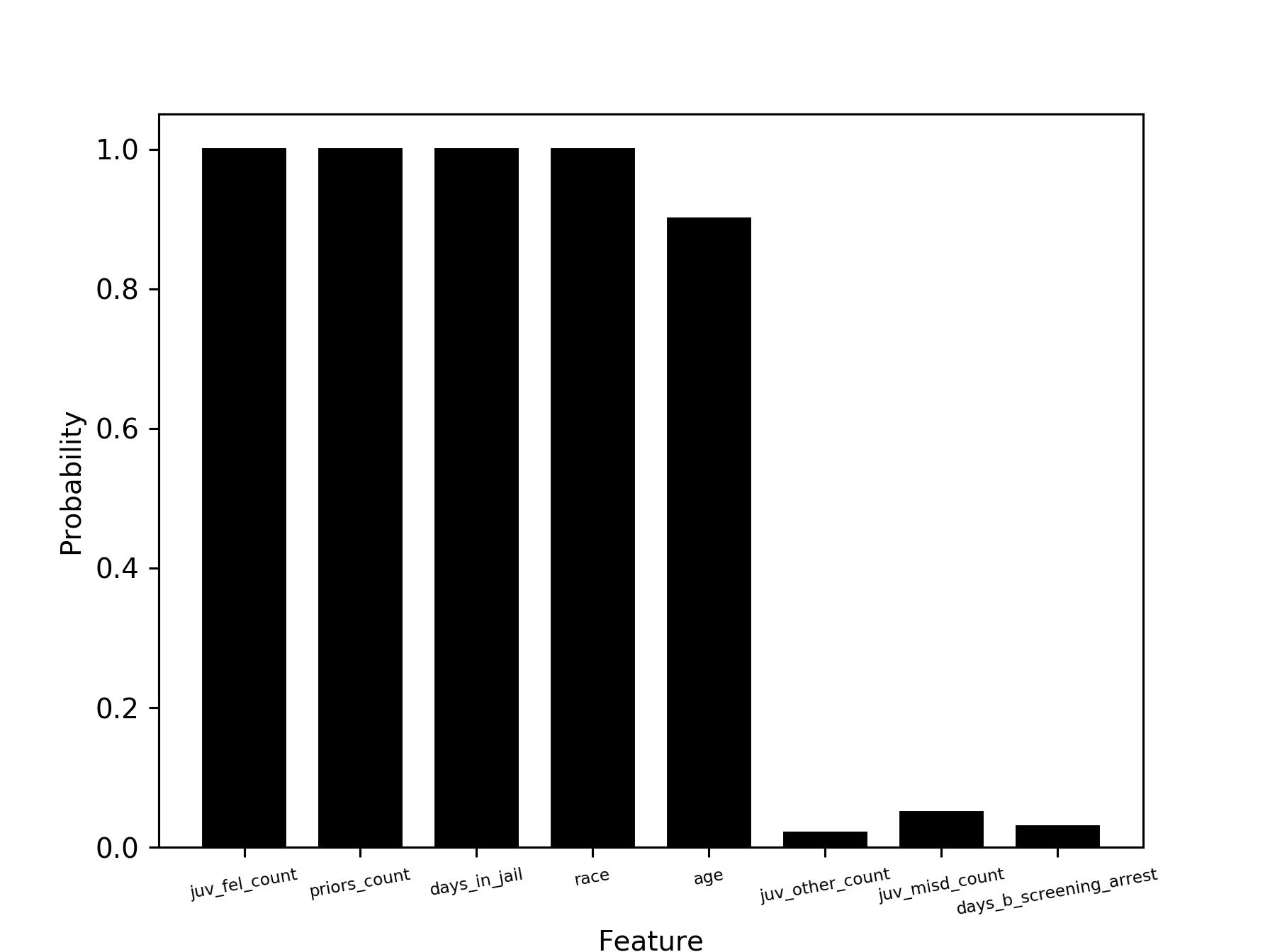}
   \caption{Sample data 2}
    \label{fig:compas-2.png}
  \end{subfigure}
 \caption{Empirical selection probability in LIME explanations of the COMPAS mimic model. LIME is run 50 times on the test points. We record five top features selected by K-LASSO and calculate the empirical selection probability for these features. The features \say{juvenile felony count}, \say{priors count}, \say{days in jail}, \say{race}, and \say{age} are consistently selected in different trials on a single data point, as well as for two different data points.}
  \label{fig:random-forest-compas}
\end{figure*}

\end{document}